\documentclass[10pt,twocolumn,letterpaper]{article}

\usepackage{cvpr}              

\usepackage{graphicx}
\usepackage{capt-of}
\usepackage{amsmath,amssymb,amsfonts,dsfont,bm,bbm,mathrsfs,pifont}
\usepackage{booktabs,multirow,adjustbox}
\usepackage[pagebackref,breaklinks,colorlinks]{hyperref}
\usepackage[capitalize]{cleveref}  
\crefname{section}{Sec.}{Secs.}
\Crefname{section}{Section}{Sections}
\crefname{table}{Tab.}{Tabs.}
\Crefname{table}{Table}{Tables}
\crefname{figure}{Fig.}{Figs.}
\Crefname{figure}{Figure}{Figures}
\crefname{equation}{Eq.}{Eqs.}
\Crefname{equation}{Equation}{Equations}
\newcommand{\R}{\mathbb{R}}       
\newcommand{\E}{\mathbb{E}}       

\newcommand{\x}{\mathbf{x}}         
\renewcommand{\d}{\mathbf{d}}       
\renewcommand{\r}{\mathbf{r}}       
\renewcommand{\c}{\mathbf{c}}       
\newcommand{\sig}{\mathbf{\sigma}}  
\newcommand{\f}{\mathbf{f}}         
\newcommand{\I}{\mathbf{I}}         
\renewcommand{\v}{\mathbf{v}}       
\newcommand{\V}{\mathbf{V}}         
\newcommand{\m}{\mathbf{m}}         
\newcommand{\M}{\mathbf{M}}         
\newcommand{\z}{\mathbf{z}}         

\def\confName{CVPR}
\def\confYear{2022}

\begin{document}

\title{3D-aware Image Synthesis via Learning Structural and Textural Representations}


\author{Yinghao Xu$^{1}$  \quad Sida Peng$^{2}$ \quad Ceyuan Yang$^{1}$ \quad Yujun Shen$^{3}$ \quad Bolei Zhou$^{1}$ \\
	$^1$The Chinese University of Hong Kong \quad
    $^2$Zhejiang University  \quad 
    $^3$Bytedance Inc. \\
    {\tt\small \{xy119, yc019, bzhou\}@ie.cuhk.edu.hk \quad    pengsida@zju.edu.cn} \quad 
	{\tt\small shenyujun0302@gmail.com} 
	}
	\vspace{-5pt}

\maketitle

\begin{abstract}
Making generative models 3D-aware bridges the 2D image space and the 3D physical world yet remains challenging.
Recent attempts equip a Generative Adversarial Network (GAN) with a Neural Radiance Field (NeRF), which maps 3D coordinates to pixel values, as a 3D prior.
However, the implicit function in NeRF has a very local receptive field, making the generator hard to become aware of the global structure.
Meanwhile, NeRF is built on volume rendering which can be too costly to produce high-resolution results, increasing the optimization difficulty.
To alleviate these two problems, we propose a novel framework, termed as \textbf{VolumeGAN}, for high-fidelity 3D-aware image synthesis, through explicitly learning a structural representation and a textural representation.
We first learn a feature volume to represent the underlying structure, which is then converted to a feature field using a NeRF-like model.
The feature field is further accumulated into a 2D feature map as the textural representation, followed by a neural renderer for appearance synthesis.
Such a design enables independent control of the shape and the appearance.
Project page is at \url{https://genforce.github.io/volumegan}.
\end{abstract}

\section{Introduction}\label{sec:intro}

Learning 3D-aware image synthesis draws wide attention recently~\cite{graf, giraffe, pigan}.
An emerging solution is to integrate a Neural Radiance Field (NeRF)~\cite{nerf} into a Generative Adversarial Network (GAN)~\cite{gan}.
Specifically, the 2D Convolutional Neural Network (CNN) based generator is replaced with a generative implicit function, which maps the raw 3D coordinates to point-wise densities and colors conditioned on the given latent code.
Such an implicit function encodes the structure and the texture of the output image in the 3D space.

However, there are two problems of directly employing NeRF~\cite{nerf} in the generator.
On one hand, the implicit function in NeRF produces the color and density for each 3D point using a Multi-Layer Perceptron (MLP) network.
With a very local receptive field, it is hard for the MLP to represent the underlying structure globally when synthesizing images.
Thus only using the 3D coordinates as the inputs~\cite{graf, pigan, giraffe} is not expressive enough to guide the generator with the global structure.
On the other hand, volume rendering generates the pixel values of the output image separately, which requires sampling numerous points along the camera ray regarding each pixel.
The computational cost hence significantly increases when the image size becomes larger.
It may cause the insufficient optimization of the model training, and further lead to unsatisfying performance for high-resolution image generation.

\begin{figure}[t]
    \centering
    \includegraphics[width=1.0\linewidth]{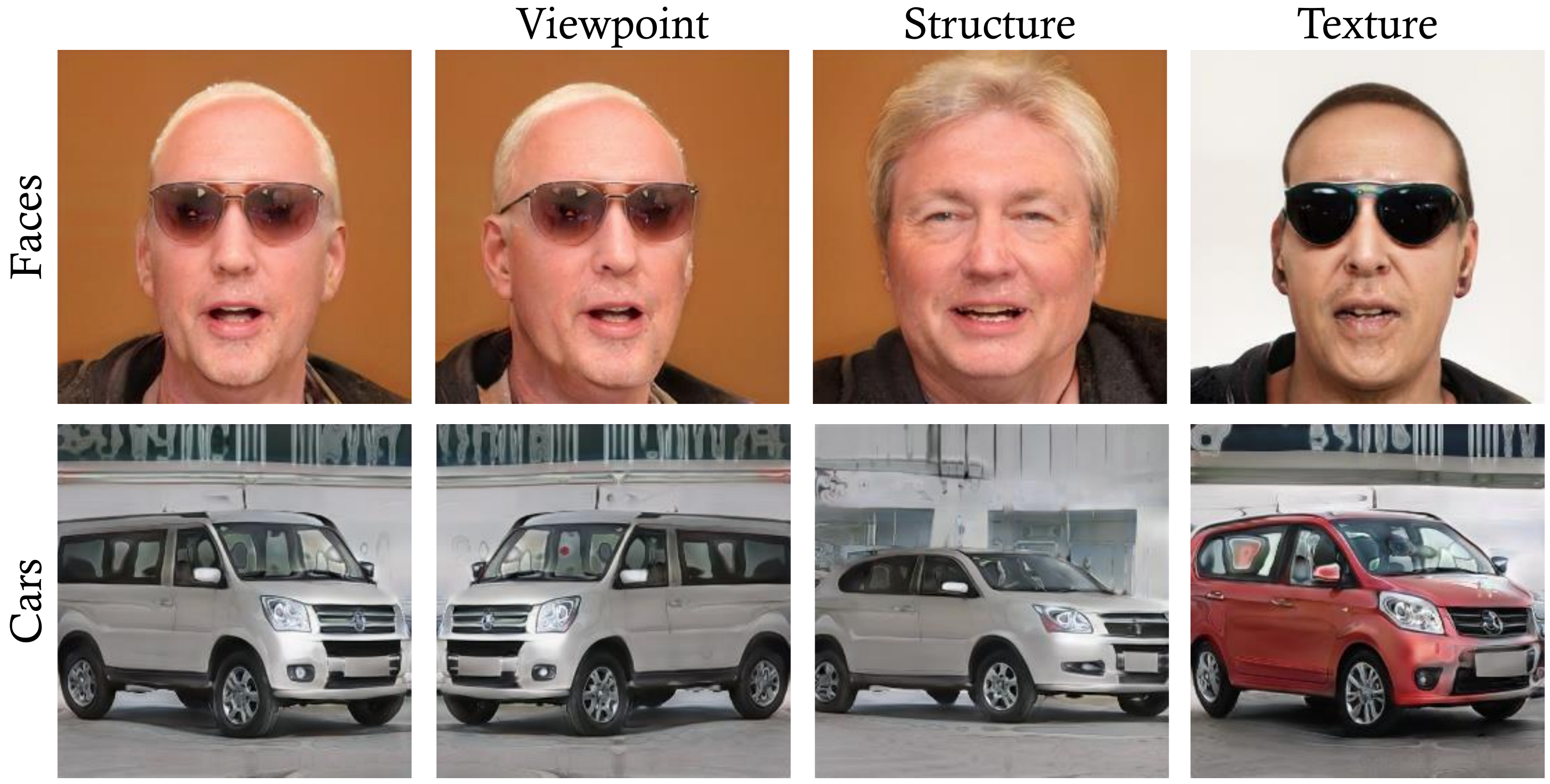}
    \vspace{-10pt}
    \caption{
        \textbf{Images of faces and cars synthesized by VolumeGAN}, which enables the control of viewpoint, structure, and texture.}
    \label{fig:teaser}
    \vspace{-20pt}
\end{figure}

Prior work has found that 2D GANs benefits from valid representations learned by the generator~\cite{interfacegan, higan, xu2021generative}.
Such generative representations describe a synthesis with \textit{high-level features}.
For example, Xu~\textit{et al.}~\cite{xu2021generative} confirm that a face synthesis model is aware of the landmark positions of the output face, and Yang~\textit{et al.}~\cite{higan} identify the multi-level variation factors emerging from generating bedroom images.
These representative features encode rich texture and structure information, thereby enhancing the synthesis quality~\cite{stylegan} and the controllability~\cite{interfacegan} of image GANs.
In contrast, as mentioned above, existing 3D-aware generative models directly render the pixel values from coordinates~\cite{graf, pigan}, without learning explicit representations.

In this work, we propose a new generative model, termed as \textbf{VolumeGAN}, which achieves 3D-aware image synthesis through \textit{explicitly learning a structural and a textural representation}.
Instead of using the 3D coordinates as the inputs, we generate a feature volume using a 3D convolutional network, which encodes the relationship between various spatial regions and hence compensates for the insufficient receptive field caused by the MLP in NeRF.
With the feature volume modeling the underlying structure, we query a coordinate descriptor from the feature volume to describe the structural information for each 3D point.
We then employ a NeRF-like model to create a feature field, by taking the coordinate descriptor attached with the raw coordinate as the input.
The feature field is further accumulated into a 2D feature map as the textural representation, followed by a CNN with $1\times1$ kernel size to finally render the output image.
In this way, we separately model the structure and the texture with the 3D feature volume and the 2D feature map, enabling the disentangled control of the shape and the appearance.

We evaluate our approach on various datasets and demonstrate its superior performance over existing alternatives.
In terms of the image quality, VolumeGAN achieves substantially better Fréchet Inception Distance (FID) score~\cite{fid}.
Taking the FFHQ dataset~\cite{stylegan} under $256\times256$ resolution as an instance, we improve the FID from $36.7$ to $9.1$.
We also enable 3D-aware image synthesis on the challenging indoor scene dataset, \textit{i.e.}, LSUN bedroom~\cite{lsun}.
Our model also suggests stable control of the object pose and shows better consistency across different viewpoints, benefiting from the learned structural representation (\textit{i.e.}, the feature volume).
Furthermore, we conduct a detailed empirical study on the learned structural and textural representations, and analyze the trade-off between the image quality and the 3D property.

\begin{figure*}[t]
    \centering
    \includegraphics[width=0.95\linewidth]{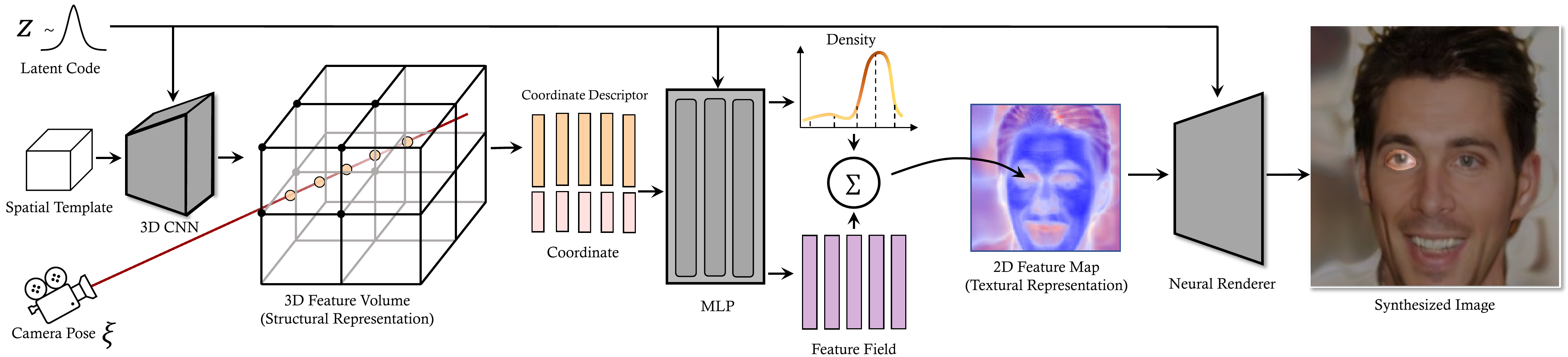}
    \caption{
        \textbf{Framework of the proposed VolumeGAN.}
        We first learn a feature volume, starting from a learnable spatial template, as the \textit{structural representation}.
        Given the camera pose $\xi$, we sample points along a camera ray and query the coordinate descriptor of each point from the feature volume via trilinear interpolation.
        The resulting coordinate descriptors, concatenated with the raw 3D coordinates, are then converted to a generative feature field and further accumulated as a 2D feature map.
        Such a feature map is regarded as the \textit{textural representation}, which guides the rendering of the appearance of the output synthesis with the help of a neural renderer.
    }
    \label{fig:framework}
    \vspace{-10pt}
\end{figure*}

\section{Related work}

\noindent{\textbf{Neural Implicit Representations.}} Recent methods \cite{sitzmann2019scene, occupancy, deepsdf, chibane2020implicit, nerf, liu2020neural} propose to represent 3D scenes with neural implicit functions, such as occupancy field \cite{occupancy}, signed distance field \cite{deepsdf}, and radiance field \cite{nerf}. To recover these representations from images, they develop differentiable renderers \cite{liu2020dist, niemeyer2020differentiable, yariv2020multiview, wang2021neus} that render implicit functions into images, and optimize the network parameters by minimizing the difference between rendered images and observed images. 
These methods can reconstruct high-quality 3D shapes and perform photo-realistic view synthesis, but they have several strong assumptions on the input data, including dense camera views, precise camera parameters, and constant lighting effects.
More recently, some methods \cite{martinbrualla2020nerfw, jain2021putting, meng2021gnerf, graf, pigan, giraffe} have attempted to reduce the constraints on the input data. 
By appending an appearance embedding to each input image, \cite{martinbrualla2020nerfw} can recover 3D scenes from multi-view images with different lighting effects. 
\cite{jain2021putting, meng2021gnerf} reconstructs neural radiance fields from very sparse views by applying a discriminator to supervise the synthesized images on novel views. 
Different from these methods requiring multi-view images, our approach can synthesize high-resolution images by training networks only on unstructured single-view image collections.

\noindent{\textbf{Image Synthesis with 2D GANs.}}
Generative Adversarial Networks (GANs)~\cite{pggan, gan} have made significant progress in synthesizing photo-realistic images but lack the ability to control the generation.
To obtain better controllability in synthesizing process, \cite{interfacegan, higan, sefa, lowrankgan} investigate the latent space of the pre-trained GANs to determine the semantic direction.
Many works~\cite{chen2016infogan, peebles2020hessian} add regularizers or modify the network structure~\cite{he2021eigengan, stylegan, stylegan2, aliasfreegan} to improve the disentanglement of variation factors without explicit supervision.
Besides, recent methods~\cite{idinvert, xu2021generative, mganprior, image2stylegan} adopt optimization or train encoders for controlling attributes of real images by pre-trained GANs.
However, these efforts control the generation only in 2D space and ignore the 3D nature of the physical world, resulting in a lack of consistency for view synthesis.

\noindent{\textbf{3D-Aware Image Synthesis.}}
2D GANs lack knowledge of 3D structure.
Some prior works directly introduce 3D representation to perform 3D-aware image synthesis.
VON~\cite{VON} generates a 3D shape represented by voxels which is then projected into 2D image space by a differentiable renderer.
HoloGAN~\cite{nguyen2019hologan} propose voxelized and implicit 3D representations and then render it to 2D space with a reshape operation.
While these methods can achieve good results, the synthesized images suffer from the fine details and identity shift because of the voxel resolution restriction.
Instead of voxel representation, GRAF~\cite{graf} and $\pi$-GAN~\cite{pigan} propose to model 3D shapes by neural implicit representation, which maps the coordinates to the RGB color.
GOF~\cite{gof} and ShadeGAN~\cite{shadegan} introduce the occupancy field and albedo field instead of radiance field for image rendering.
However, due to the computationally intensive rendering process, they cannot synthesize high-resolution images with good visual quality.
To overcome this problem, \cite{giraffe} first render low-resolution feature maps with neural feature fields and then generate high-resolution images with 2D CNNs, also with the coordinates as the input.
However, severe artifacts across different camera views are introduced because CNN-based decoder harms the 3D consistency.
Unlike previous attempts, we leverage the feature volume to provide the feature descriptor for each coordinate and a neural renderer consisting of $1\times1$ convolution block to synthesize high-quality images with better multi-view consistency and 3D control.
The concurrent work StyleNeRF~\cite{gu2021stylenerf} also adopts $1\times1$ convolution block to synthesize high-quality images.
However, we adopt the feature volume to provide the structural description for the synthesized object instead of using regularizers to improve the 3D properties. 
\section{Method}\label{sec-method}

This work targets at learning 3D-aware image generative models from a collection of 2D images.
Previous attempts replace the generator of a GAN model with an implicit function~\cite{nerf}, which maps 3D coordinates to pixel values.
To improve the controllability and synthesis quality, we propose to explicitly learn the structural and the textural representations that are responsible for the underlying structure and texture of the object respectively.
Concretely, instead of directly bridging coordinates with densities and RGB colors, we ask the implicit function to transform 3D feature volume (\textit{i.e.}, the structural representation) to a generative feature field, which are then accumulated into a 2D feature map (\textit{i.e.}, the textural representation).
The overall framework is illustrated in \cref{fig:framework}.
Before going into details, we first briefly introduce the Neural Radiance Field (NeRF), which is a core module of the proposed model.

\subsection{Preliminary}\label{sec-method-preliminary}

The neural radiance field~\cite{nerf} is formulated as a continuous function $F(\x, \d) = (\c, \sig)$, which maps a 3D coordinate $\x\in\R^3$ and the viewing direction $\d\in\mathbb{S}^2$ to the RGB color $\c\in\R^3$ and a volume density $\sig\in\R$.
Then, given a sampled ray, we can predict the colors and densities of all the points that the ray goes through, which are then accumulated into the pixel value with volume rendering techniques.
Typically, the function $F(\cdot, \cdot)$ is parameterized with a multi-layer perceptron (MLP), $\Phi(\cdot)$, as the backbone, and two independent heads, $\phi_c(\cdot, \cdot)$ and $\phi_d(\cdot)$, to regress the color and density:
\begin{align}
    \c(\x, \d) &= \phi_c(\Phi(\x), \d), \label{eq:nerf-color} \\
    \sig(\x) &= \phi_d(\Phi(\x)), \label{eq:nerf-density}
\end{align}
where the color is related with the viewing direction $\d$ due to the variation factors like lighting, while the density $\sigma$ is independent of $\d$.

NeRF is primarily proposed for 3D reconstruction and novel view synthesis, which is trained with the supervision from multi-view images.
To enable random sampling by learning from a collection of single-view images, recent attempts~\cite{graf, pigan} introduce a latent code $\z$ to the function $F(\cdot, \cdot)$.
In this way, the geometry and appearance of the rendered image will vary according to the input $\z$, resulting in diverse generation.
Such a stochastic implicit function is asked to compete with a discriminator of GANs~\cite{gan} to mimic the distribution of real 2D images.
In the learning process, the revised function $F(\x, \d, \z) = (\c, \sig)$ is supposed to encode the structure and the texture information simultaneously.

\subsection{3D-aware Generator in VolumeGAN} \label{sec-method-volumegan}

To improve the controllability and image quality of the NeRF-based 3D-aware generative model, we propose to explicitly learn a structural representation and a textural representation, which control the underlying structure and texture respectively.
In this part, we will introduce the design of the structural and the textural representations, as well as their integration through a generative neural feature field.

\vspace{5pt}
\noindent\textbf{3D Feature Volume as Structural Representation.}
As pointed out by NeRF~\cite{nerf}, the low-dimensional coordinates $\x$ should be projected into a higher-dimensional feature to describe the complex 3D scenes.
For this purpose, a typical solution is to characterize $\x$ into Fourier features~\cite{vaswani2017attention}.
However, such a Fourier transformation cannot introduce additional information beyond the spatial position. It may be enough for reconstructing a fixed scene, but yet far from encoding a distributed feature for the image synthesis of different object instances.
Hence, we propose to learn a grid of features providing the inputs of implicit functions, which gives a more detailed description of each spatial point.
We term such a 3D feature volume, $\V$, as the \textit{structural representation} which characterizes the underlying 3D structure.
To obtain the feature volume, we employ a sequence of 3D convolutional layers with the Leaky ReLU (LReLU) functions~\cite{lrelu}.
Inspired by Karras~\textit{et al.}~\cite{stylegan}, we apply Adaptive Instance Normalization (AdaIN)~\cite{adain} to the output of each layer to introduce diversity to the feature volume.
Starting from a learnable 3D tensor, $\V_0$, the structural representation is generated with
\begin{align}
     \V &= \psi_{n_s - 1} \circ \psi_{n_s - 2} \circ ... \circ \psi_0({\V_0}), \\
     \psi_i(\V_i) &= \mathtt{AdaIN}\Big(\mathtt{LReLU}\big(\mathtt{Conv}(\mathtt{Up}(\V_i, s_i))\big), \z \Big),
\end{align}
where $n_s$ denotes the number of layers for structure learning.
$s_i \in \{1, 2\}$ is the upsampling scale for the $i$-th layer.

\vspace{5pt}
\noindent\textbf{2D Feature Map as Textural Representation.}
As discussed before, volume rendering can be extremely slow and computationally expensive, making it costly to directly render the raw pixels of a high-resolution image.
To mitigate the issue, we propose to learn a feature map at a low resolution, followed by a CNN to render a high-fidelity result.
Here, the 2D feature map is responsible for describing the visual appearance of the final output.
The tailing CNN consists of several Modulated Convolutional Layers (ModConv)~\cite{stylegan2}, also activated by LReLU.
To avoid the CNN from weakening the 3D consistency~\cite{giraffe}, we use $1\times1$ kernel size for all layers such that the per-pixel feature can be processed independently.
In particular, given a 2D feature map, $\M$, as the textural representation, the image is generated by
\begin{align}
     \I^f &= f_{n_t - 1} \circ f_{n_t - 2} \circ ... \circ f_0(\M), \\
     f_i(\M_i) &= \mathtt{LReLU}\big(\mathtt{ModConv}(\M_i, t_i, \z) \big),
\end{align}
where $n_t$ denotes the number of layers for texture learning.
$t_i \in \{1, 2\}$ is the upsampling scale for the $i$-th layer.

\vspace{5pt}
\noindent\textbf{Bridging Representations with Neural Feature Field.}
To connect the structural and the textural representations in the framework, we introduce a neural radiance field~\cite{nerf} as the bridge.
Different from the implicit function in the original NeRF, which maps coordinates to pixel values, we first query the coordinate descriptor, $\v$, from the feature volume, $\V$, given a 3D coordinate $\x$, and then concatenate it with $\x$ to obtain $\v^{\x}$ as the input.
Then, the implicit function transform $\v^{\x}$ to the density and feature vector of the field.
The above process can be formulated as
\begin{align}
    \v &= \mathtt{trilinear}(\V, \x), \\
    \v^{\x} &= \mathtt{Concat}(\v, \x), \label{eq:coord-concat} \\
    \Phi(\v^{\x}) &= \phi_{n-1} \circ \phi_{n-2} \circ ... \circ \phi_0(\v^{\x}), \\
    \phi_i(\v^{\x}_i) &= \sin\big(\gamma_i(\z)\cdot (\mathbf{W}_i \v^{\x}_i + \mathbf{b}_i) + \beta_i(\z))\big), \label{eq:siren} \\
    \f(\x, \d) &= \phi_f(\Phi(\v^{\x}), \d), \label{eq:feature-field} \\
    \sigma(\x) &= \phi_d(\Phi(\v^{\x})), 
\end{align}
where $n$ denotes the number of layers to parameterize the neural field, while $\mathbf{W}_i$ and $\mathbf{b}_i$ are the learnable layer-wise weight and bias.
\cref{eq:coord-concat} concatenates coordinates $\x$ onto feature $\v$ to explicitly introduce the structural information.
\cref{eq:siren} follows Chan~\textit{et al.}~\cite{pigan}, which conditions the layer-wise output of the backbone $\Phi(\cdot)$ on the frequencies, $\gamma_i(\cdot)$, and phase shifts, $\beta_i(\cdot)$, learned from the random noise $\z$.
\cref{eq:feature-field} replaces the color modeling in \cref{eq:nerf-color} with feature modeling.

A per-pixel final feature $\m$ can be obtained via volume rendering along a ray $\r$ (with viewing direction $\d$). 
A collection of $\m$ regarding different rays group into a 2D feature map as the \textit{textural representation}, $\M$, which will be further used to render the image.
\vspace{-5pt}
\begin{align}
        \m(\r) &= \sum_{k=1}^{N} T_k (1 - \exp(-\sig(\x_k) \delta_k)) \f(\x_k, \d), \label{eq:integral}\\
    T_k &= \exp(-\sum_{j=1}^{k-1} \sigma(\x_j) \delta_j).
\end{align}
%
\cref{eq:integral} approximates the integral of $N$ points $\{\x_k\}_{k=1}^{N}$ on the sampled ray $\r$, where $\delta_k = || \x_{k + 1} - \x_{k} ||_2$ stands for the distance between adjacent sampled points.

\subsection{Training Pipeline} \label{sec-method-training}

\noindent\textbf{Generative Sampling.}
The whole generation process is formulated as $\I^f=G(\z, \xi)$, where $\z$ is a latent code sampled from a Gaussian distribution $\mathcal{N}(0, 1)$ and $\xi$ denotes the camera pose sampled from a prior distribution $p_\xi$. 
$p_\xi$ is tuned for different datasets as either Gaussian or Uniform.

\vspace{2pt}
\noindent\textbf{Discriminator.}
Like existing approaches for 3D-aware image synthesis~\cite{graf, pigan, giraffe}, we employ a discriminator $D(\cdot)$ to compete with the generator.
The discriminator is a CNN consisting of several residual blocks like~\cite{stylegan2}.

\begin{figure*}[t]
    \centering
    \includegraphics[width=1.0\textwidth]{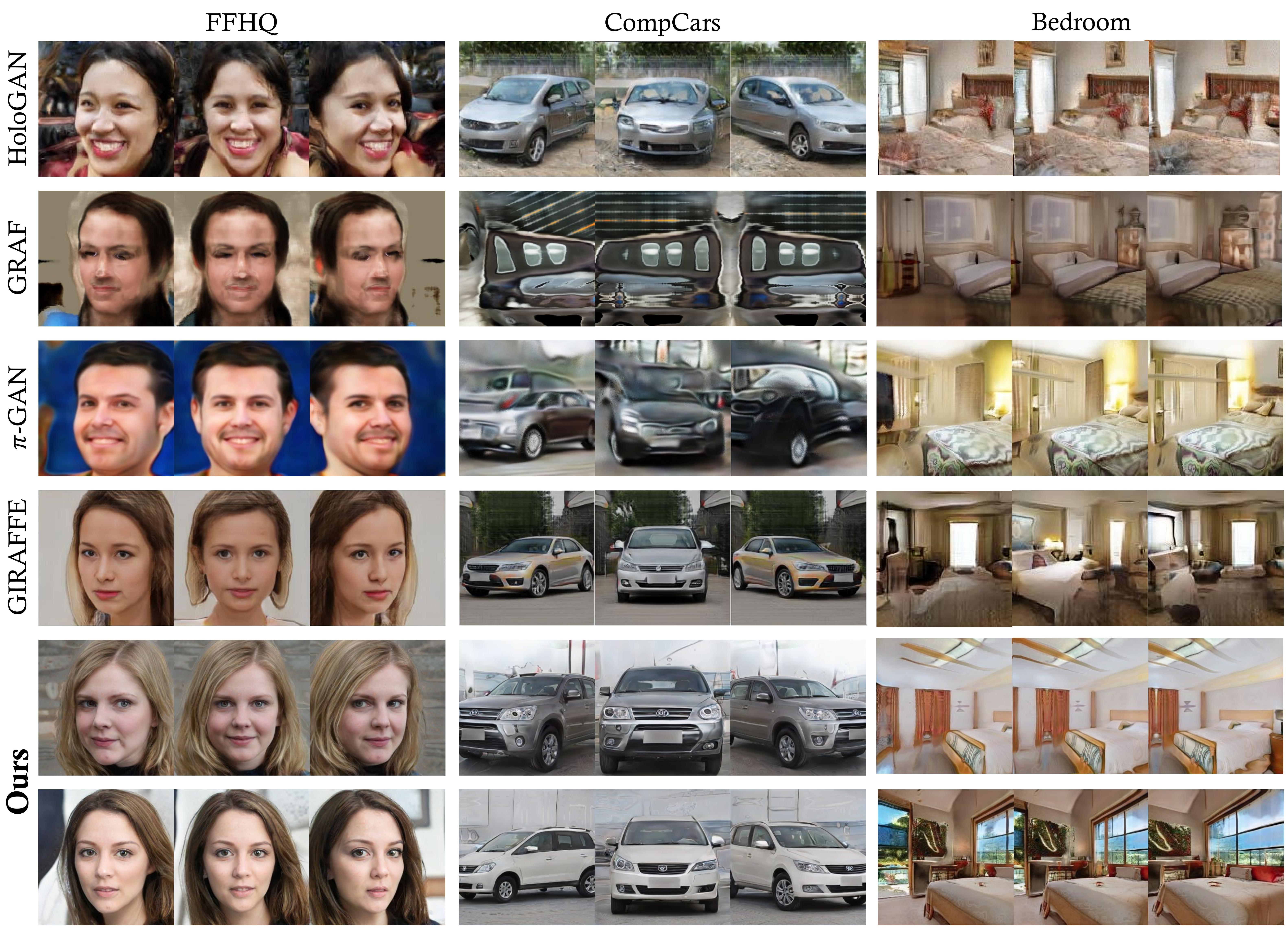}
    \vspace{-20pt}
    \caption{
        \textbf{Qualitative comparison between our VolumeGAN and existing alternatives} on FFHQ~\cite{stylegan}, CompCars~\cite{compcars}, and LSUN bedroom~\cite{lsun} datasets. All images are in $256\times 256$ resolution.}
    \label{fig:comparison}
    \vspace{-15pt}
\end{figure*}

\vspace{2pt}
\noindent\textbf{Training Objectives.}
During training, we randomly sample $\z$ and $\xi$ from the prior distributions, while the real images $\I^r$ are sampled from the real data distribution $p_\mathcal{D}$. The generator and the discriminator are jointly trained with
\begin{align}
    \min\mathcal{L}_G &= \E_{\z \sim p_z, \xi \sim p_\xi}[f(D(G(\z, \xi)))], \\
    \min\mathcal{L}_D &= \E_{\I^r \sim p_D}[f(-D(\I^r)) +
                         \lambda ||\nabla_{\I^r}D(\I^r)||_2^2] \label{eq:d_loss},
\end{align}
where $f(t) = -\log(1+\exp(-t)$ is the softplus function. The last term in Eq.~\eqref{eq:d_loss} stands for the gradient penalty regularizer and $\lambda$ is the loss weight.

\section{Experiment}

\subsection{Settings}~\label{sec:exp:settings}

\vspace{-10pt}
\noindent\textbf{Datasets.}
We evaluate the proposed VolumeGAN on five real-world unstructured datasets including CelebA~\cite{celeba}, Cats~\cite{cat}, FFHQ~\cite{stylegan}, CompCars~\cite{compcars}, LSUN
bedroom~\cite{lsun}, and a synthetic dataset Carla~\cite{carla}.
CelebA contains around 20$K$ face images from 10$K$ identities. 
The crop from the top of the hair to the bottom of the chin is adopted for data preprocessing on CelebA.
The Cats dataset contains 6.5$K$ images of cat heads at $128\times 128$ resolution.
FFHQ contains 70$K$ images of real human faces in a resolution of $1024\times 1024$. 
We follow the protocol of~\cite{stylegan} to preprocess the faces of FFHQ.
Compcars includes 136$K$ real cars whose pose varies greatly. 
The original images are in different aspect ratios. 
Hence we center crop the cars and resize them into $256\times 256$.
Carla dataset contains 10$K$ images which are rendered from Carla Driving simulator~\cite{carla} using 16 car models with different textures.  
LSUN bedroom includes 300$K$ samples in various camera views and aspect ratios.
We also use center cropping to preprocess the bedroom images.
We train VolumeGAN on resolutions of  $128\times 128$ for CelebA, Cats, and Carla and  $256\times 256$ for FFHQ, CompCars and LSUN bedroom.

\noindent\textbf{Baselines.}
We choose four 3D-aware image synthesis approaches as the baselines, including HoloGAN~\cite{nguyen2019hologan}, GRAF~\cite{graf}, $\pi$-GAN~\cite{pigan} and GIRAFFE~\cite{giraffe}. 
Baseline models are officially released by the original papers or trained with the official implementation.
More details can be found in \textit{Supplementary Material}.%
\vspace{-3pt}%
\footnote{We fail to reproduce HoloGAN on LSUN bedroom with the \href{https://github.com/thunguyenphuoc/HoloGAN}{official implementation}, hence we do not report the quantitative results. The qualitative results of bedrooms are borrowed from the original paper~\cite{nguyen2019hologan}.}

\definecolor{azure}{rgb}{0.0, 0.5, 1.0}
\begin{table*}[t]
\setlength{\tabcolsep}{7.5pt}
\caption{
    \textbf{Quantitative comparisons on different datasets.}
    FID~\cite{fid} (lower is better) is used as the evaluation metric.
    Numbers in brackets indicate the improvements of our VolumeGAN over the second method.
}
\label{tab:comparison}
\vspace{-8pt}
\centering
\begin{tabular}{l|c|c|c|c|c|c}
\toprule
\multirow{1}{*}{Method}       & \multicolumn{1}{c|}{CelebA 128} & \multicolumn{1}{c|}{Cats 128} & \multicolumn{1}{c|}{Carla 128} & \multicolumn{1}{c|}{FFHQ 256} & \multicolumn{1}{c|}{CompCars 256} & \multicolumn{1}{c}{Bedroom 256} \\ \hline
HoloGAN~\cite{nguyen2019hologan}     &   39.7 & 40.4 & 126.4 & 72.6 & 65.6    & $-$ \\
GRAF~\cite{graf}        &   41.1 & 28.9 & 41.6 &  81.3 & 222.1   & 63.9            \\
$\pi$-GAN~\cite{pigan}   &   15.9 & 17.7 & 30.1 &  53.2 & 194.5 & 33.9       \\
GIRAFFE~\cite{giraffe}    &   17.5     & 20.1    & 30.8 & 36.7 & 27.2 & 44.2 \\ \hline
VolumeGAN (Ours)        & \textbf{8.9 \textcolor{azure}{($-$7.0)} }   & \textbf{5.1 \textcolor{azure}{($-$12.6)} } & \textbf{7.9 \textcolor{azure}{($-$22.2)}} & \textbf{9.1 \textcolor{azure}{($-$27.6)}} & \textbf{12.9 \textcolor{azure}{($-$14.3)}} & \textbf{17.3 \textcolor{azure}{($-$16.6)}}   \\ \bottomrule
\end{tabular}
\vspace{-15pt}
\end{table*}

\noindent\textbf{Implementation Details.}
The learnable 3D template $\mathbf{V_0}$ are randomly initialized in $4\times 4 \times4$ shape and 
3D convolutions with kernel size $3\times 3 \times3$ are stacked to embed the template, resulting in the feature volume in $32\times 32 \times 32$ resolution. 
We sample rays in a resolution of $64\times 64$, and four conditioned MLPs (SIREN~\cite{pigan, siren}) with 256 dimensions are adopted to model the feature field and the volume density.
We use an Upsample block~\cite{stylegan2} and two $1\times1$ $\mathtt{ModConv}$~\cite{stylegan2, cips} at each resolution for the neural renderer until reaching the output image resolution. 
We also apply progressive training strategy used in PG-GAN~\cite{pggan} to achieve better image qualities.
For the network training, we use Adam~\cite{adam} optimizer with $\beta_0 = 0$ and $\beta_1 = 0.999$ over 8 GPUs. The entire training requires the discriminator to see 25000$K$ images.
The batch size is 64, and the weight decay is 0. 
%
%
More details can be found in \textit{Supplementary Material}.

\subsection{Main Results}
\noindent\textbf{Qualitative Results.} \cref{fig:comparison} compares the synthesized images of our method with baselines on FFHQ, CompCars and LSUN bedroom. 
The images are sampled from three views and synthesized in a resolution of $256\times 256$ for visualization.
Although all baseline methods can synthesize images under different camera poses on FFHQ, they suffer from low image quality and the identity shift across different angles. 
When transferred to challenging CompCars with larger viewpoint variations, some baselines such as GRAF~\cite{graf} and $\pi$-GAN~\cite{pigan} struggle to generate realistic cars.
HoloGAN can achieve good image quality but suffers from multi-view inconsistency.
GIRAFFE can generate realistic cars while the color of the cars changes significantly under different views.
When tested on bedroom, HoloGAN, GRAF, $\pi$-GAN and GIRRAFE cannot handle such indoor scene data with larger structure and texture variations. 

VolumeGAN can synthesize high-fidelity view-consistent images.
Compared with the existing approaches, it generates more fine details, such as teeth (face), headlights (car) and windows (bedroom).
Even on the more challenging CompCars and LSUN bedroom datasets, VolumeGAN still achieves satisfying synthesis performance thanks to the feature volume and the neural renderer.

\noindent\textbf{Quantitative Results.} 
We quantitatively evaluate the visual quality of the synthesized images using Frechet Inception Distance (FID)~\cite{fid}.
We follow the evaluation protocol of StyleGAN~\cite{stylegan} which adopts 50$K $ real and fake samples to calculate the FID score.
All baseline models are evaluated with the same setting for a fair comparison.
As shown in ~\cref{tab:comparison}, our approach leads to a significant improvement compared with baselines, particularly on the challenging datasets with the larger pose variation or the finer details.
Note that although GIRAFFE also uses the neural renderer, our method still outperforms it with a clear margin.
It demonstrates that the structural information encoded in the feature volume provides representative visual concepts, resulting in better images quality. 

\begin{figure}[t]
    \centering
    \includegraphics[width=0.9\linewidth]{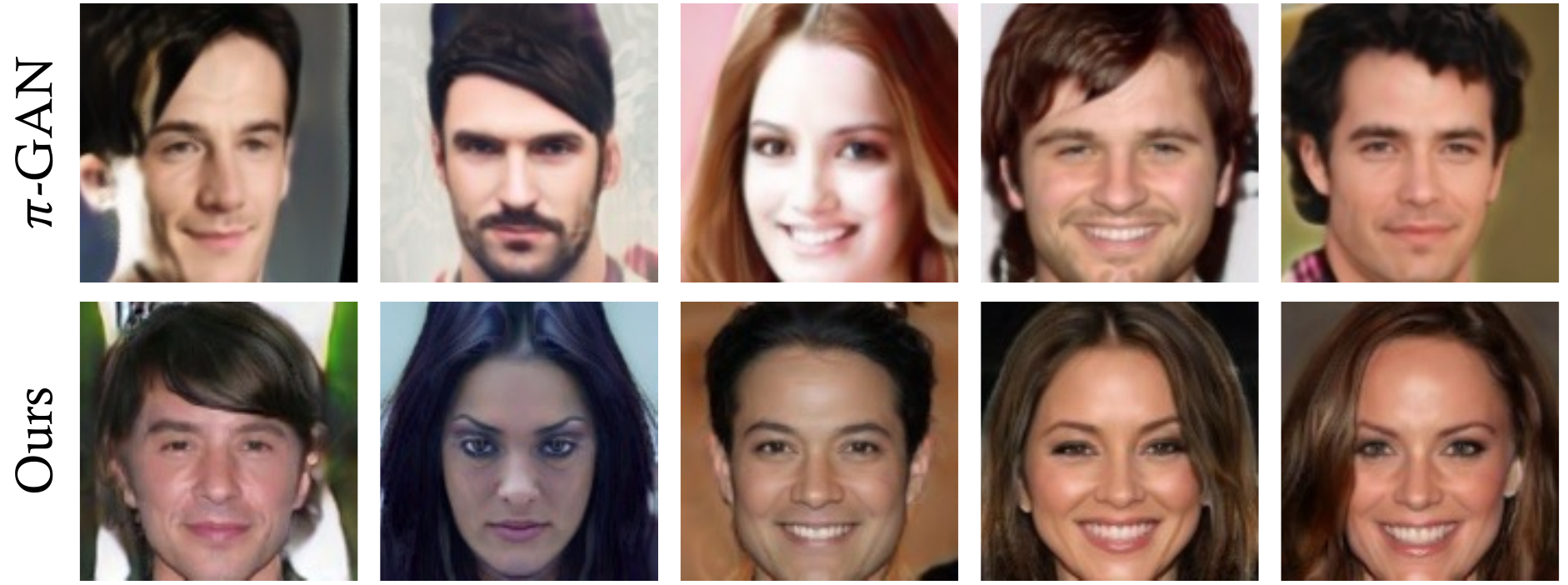}
    \vspace{-10pt}
    \caption{
        \textbf{Synthesized results with the front camera view} by $\pi$-GAN~\cite{pigan} and our VolumeGAN, where the faces proposed by VolumeGAN are more consistent to the given view, suggesting a better 3D controllability.
    }
    \label{fig:pose_error}
    \vspace{-15pt}
\end{figure}

\subsection{Ablation Studies}

We conduct ablation studies on CelebA $128\times 128$ to examine the importance of each component in VolumeGAN.

\noindent{\textbf{Metrics.}}
In addition to the \textit{FID} score that measures the image quality, we also provide two quantitative metrics to measure the multi-view consistency and the precision of 3D control as follows.
1) \textit{Reprojection Error}.
We first extract the underlying geometry of an object from the generated density using marching cubes~\cite{lorensen1987marching}.
Then, We render each object in sequence and sample five viewpoints uniformly to synthesize the images.

The depth of each image is rendered from the resulting extracted mesh, which is used to calculate the reprojection error on two consecutive views by warping them each other.
Specifically, we fix the yaw to be 0 and sample pitch from $[-0.3, 0.3]$. 
The threshold of marching cube is set to 10 due to the best visualization results of meshes.
The reprojection error is calculated in the normalized image space $[-1, +1]$ like \cite{image2stylegan, idinvert, xu2021generative} to evaluate the multi-view consistency.
2) \textit{Pose Error}. We synthesize 20,000 images and regard the results predicted from the head pose estimator~\cite{zhou2020whenet} as the ground truth.
The L1 distance between the given camera pose and the predicted pose is reported to evaluate the 3D control quantitatively.

\noindent\textbf{Ablations on VolumeGAN Components.}
Our approach proposes to use Feature Volume as the structural representation and adopt the neural renderer consisting of $\mathtt{ModConv}$ to render textural representation into high-fidelity images.
We ablate them to better understand their individual contributions.
Our baseline is built upon $\pi$-GAN~\cite{pigan} using conditioned MLPs to achieve 3D-aware image synthesis by mapping coordinates to RGB color.
The layer number of the baseline is set to be 4, the same as our setting illustrated in \cref{sec:exp:settings}  for a fair comparison.
As shown in \cref{tab:componnets}, introducing the feature volume that provides the structural representation could further improve the FID score of the baseline approach from 18.7 to 13.6.

More importantly, lower reprojection error and pose error are also achieved, demonstrating the structural representation from the feature volume not only facilitates better visual results but also maintains the 3D properties regarding multi-view consistency and 3D explicit controlling.
On top of this, the neural renderer further enhances FID to 8.9 with a slight drop in reprojection error and pose error, leading to the new state-of-the-art result on 3D-aware image synthesis.
Notably, involving the neural renderer to the baseline could also boost the FID score but apparently sacrifice the 3D properties to some extent according to the 3D metrics. It also indicates that FID is not a comprehensive metric to evaluate 3D-aware image synthesis.   
In addition, \cref{fig:pose_error} gives several synthesized samples of $\pi$-GAN baseline and our approach under the front view. 
More samples can be found in \textit{Supplementary Material}.
Qualitatively, the poses of our synthesized samples are closer to the given camera view which is quantitatively reflected by the pose-error score.

\noindent\textbf{Resolution of the Feature Volume.}
The feature volume resolution depicts the spatial refinement of the structural representation, and thus it plays an essential role in synthesizing images.
\cref{tab:voxelsize} presents the metrics of the synthesis results for various resolutions of feature volume.
As the resolution increases, the multi-view consistency and 3D control become better consistently while the visual quality measured by FID fluctuates little.
This demonstrates that a more detailed feature volume provides better geometry consistency across various camera poses.
However, increasing the feature volume resolution inevitably results in a greater computational burden. 
As a result, we choose a feature volume resolution of 32 in all of our experiments to maintain the balance between efficiency and image quality.

\noindent\textbf{Neural Renderer Depth.}
The neural renderer is adopted to convert textural representations into 2D images; thus, its capacity is critical to the quality of the generated images.
We adjust its capacity by varying the depth of the neural renderer to investigate its effect.
\cref{tab:rendererdepth} shows a trade-off between image quality and 3D properties.
As the depth of the network increases, better image visual quality can be achieved while the quality of multi-view consistency and 3D control downgrades.
This implies that increasing the capacity of the neural renderer would damage the 3D structure to some extent, revealing FID is not a comprehensive metric for 3D-aware image synthesis again. 
We thus choose the shallower network as the neural renderer for better 3D consistency and control.

\begin{table}[t]
    \centering
     \setlength{\tabcolsep}{12.5pt}
      \caption{\textbf{Ablation studies on the components of VolumeGAN}, including the feature volume (FV) and the neural renderer (NR).
    ``Rep-Er'' and ``Pose-Er'' are the reprojection-error and pose-error.}
    \vspace{-8pt}
    \begin{tabular}{cc|ccc}
    \toprule
      FV       & NR   & FID         & Rep-Er        & Pose-Er   \\ \hline
        \multicolumn{2}{c|}{$\pi$-GAN}                     & 18.7          & 0.071            & 12.7    \\ 
            \ding{51}  &            & 13.6          &  \textbf{0.031}         &   \textbf{8.3} \\
                       & \ding{51}  & 11.3        &  0.103          & 12.1   \\
    \ding{51}          & \ding{51}  & \textbf{8.9}  &   0.037         &   8.6 \\ \bottomrule
    \end{tabular}
    \label{tab:componnets}
    \vspace{-5pt}
\end{table}

\begin{table}[t]
    \centering
     \setlength{\tabcolsep}{8pt}
    \caption{\textbf{Effect of the size of feature volume.} ``Str Res'' denotes the resolution of the feature volume (\textit{i.e.}, the structural representation).}
    \vspace{-8pt}
    \begin{tabular}{c|ccc|c}
    \toprule
    Str Res   & FID     & Rep-Er        & Pose-Er & Speed (fps) \\ \hline
             16 & 9.0  & 0.040      &   9.1   &   5.58 \\ 
             32 & \textbf{8.9}  & 0.037   &   8.6  & 5.15\\
             64 & 9.2  &  \textbf{0.032}       &   \textbf{8.4} & 3.86
      \\ \bottomrule
    \end{tabular}

    \label{tab:voxelsize}
    \vspace{-5pt}
\end{table}

\begin{table}[t]
    \centering
     \setlength{\tabcolsep}{10pt}
    \caption{\textbf{Effect of the depth of neural renderer.} ``Tex Res'' denotes the resolution of the 2D feature map (\textit{i.e.}, the textural representation).}
    \vspace{-8pt}
    \begin{tabular}{cc|ccc}
    \toprule
    Depth & Tex Res  & FID     & Rep-Er        & Pose-Er   \\ \hline
      6    &   64 &  \textbf{8.0}    &  0.051          &  9.7   \\ 
      4    &   64 &  8.8    &  0.046           &   9.3   \\
      2    &   64 &  8.9 & \textbf{0.037}   &   \textbf{8.6} 
      \\ \bottomrule
    \end{tabular}
    \label{tab:rendererdepth}
    \vspace{-10pt}
\end{table}

\begin{figure*}[t!]
    \centering
    \includegraphics[width=0.95\textwidth]{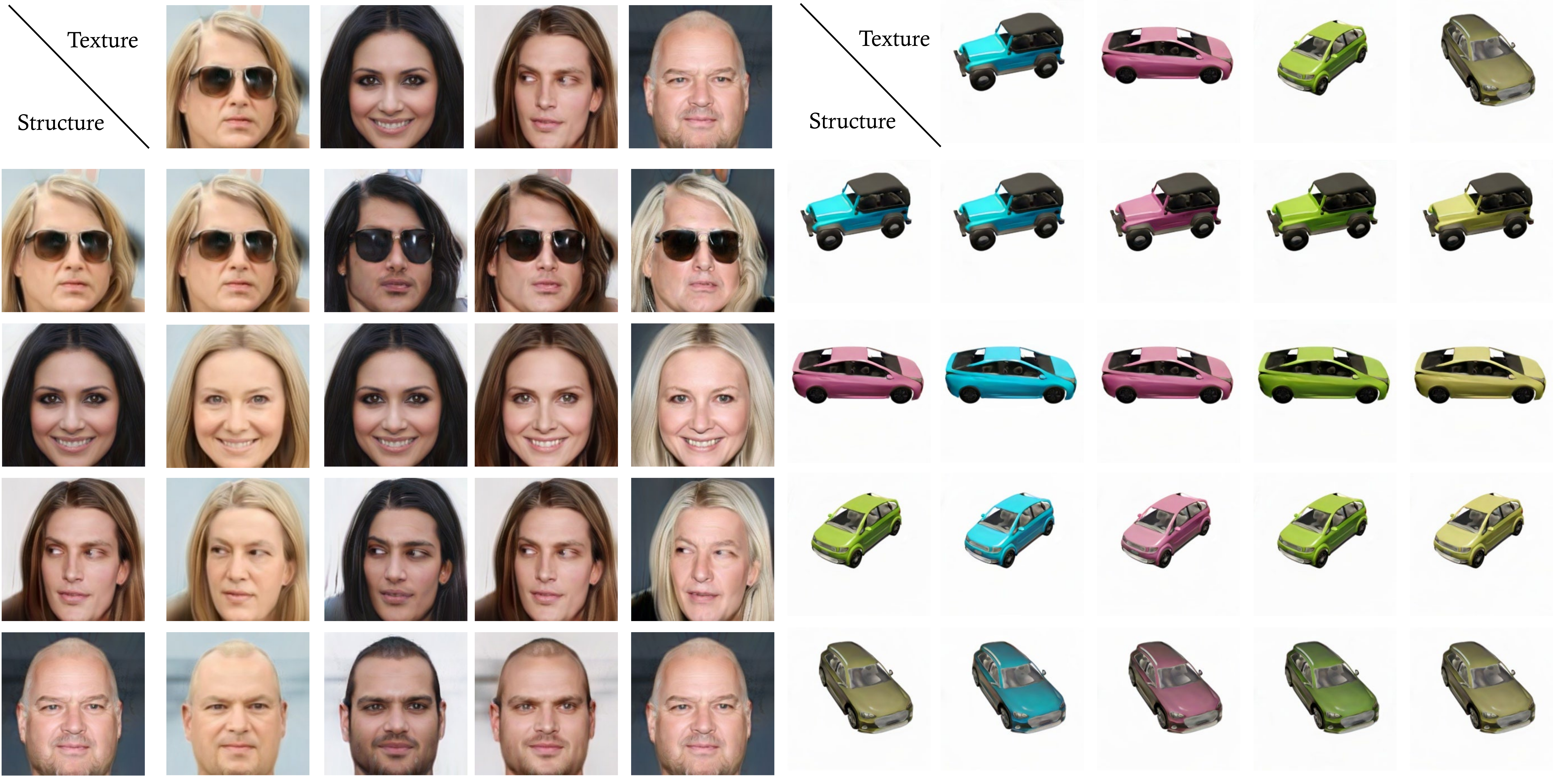}
    \vspace{-10pt}
    \caption{
        \textbf{Synthesized results by exchanging the structural and the textural latent codes.}
    }
    \label{fig:stylemixing}
    \vspace{-15pt}
\end{figure*}

\subsection{Properties of Learned Representations}

A key advantage of our approach over previous attempts is that by separately modeling the structure and texture with the 3D feature volume and 2D feature map, our model learns the disentangle representations for the object. These representations allow us to achieve control of the shape and appearance. The coordinate descriptor and the 3D mesh extracted from the density are visualized to interpret the learned representations.

\noindent{\textbf{Independent Control of Structure and Texture.}}
At test time, we could easily swap and combine the latent codes regarding the structural and textural individually.
In this way, we can investigate whether such two representations are well disentangled.
For example, we could combine the structural representation (\emph{i.e.}, feature volume code) of a certain instance with the textural (\emph{i.e.}, generative feature field and neural renderer code) of another. The corresponding results are shown in Fig.~\ref{fig:stylemixing}.
The faces results show that the feature volume code controls the shape of the face and hairstyle, whereas the feature field and neural renderer code determine the skin and hair color. 
Concretely, glasses are controlled by the volume code, in line with our perception.
We can swap the structure and texture of cars successfully.
It demonstrates that our method can disentangle shape and appearance in synthesizing images.
Different from GRAF~\cite{graf} and GIRRAFE~\cite{giraffe}, we do not explicitly introduce shape code and appearance code to control image synthesis.
Thanks to the structural and textural representations in our framework, the disentanglement between shape and appearance emerges naturally.

\begin{figure}[t]
    \centering
    \includegraphics[width=0.9\linewidth]{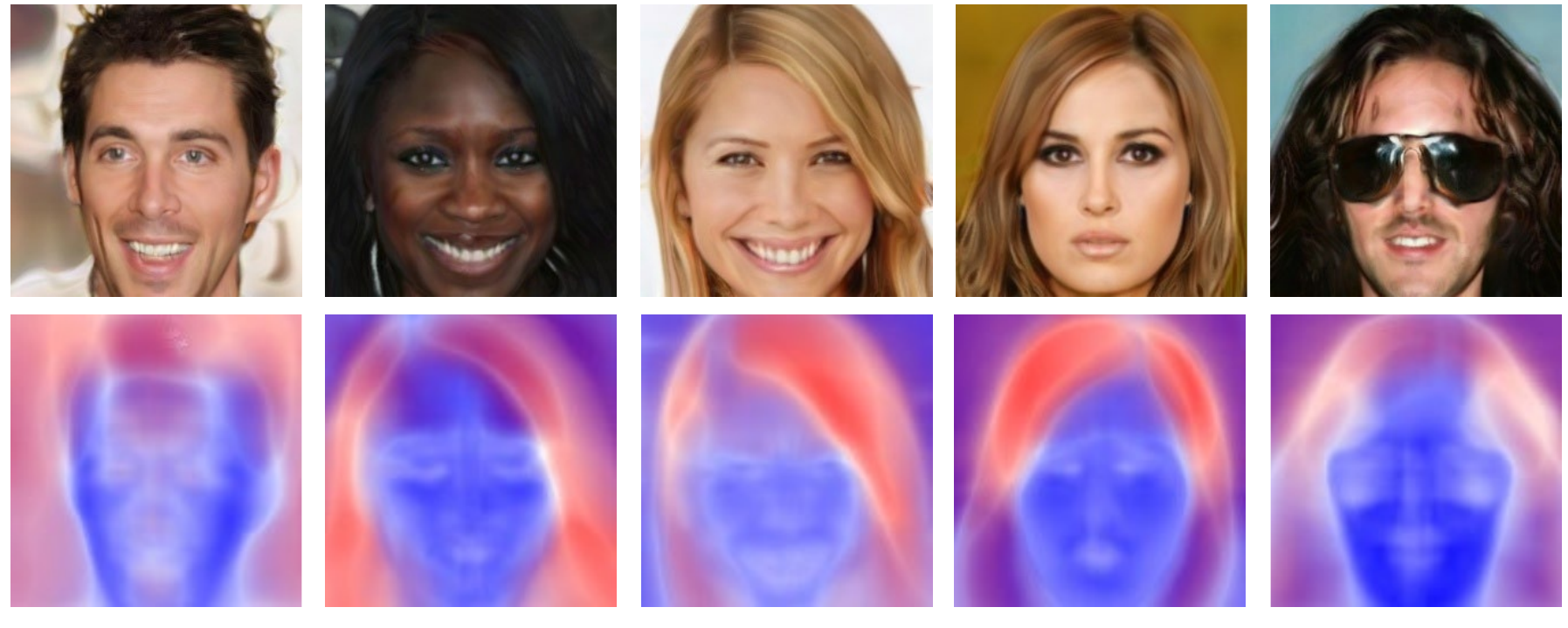}
    \vspace{-10pt}
    \caption{
        \textbf{Visualization of coordinate descriptor.}
        PCA is used to reduce the feature dimension.
    }
    \vspace{-20pt}
    \label{fig:featurevis}
\end{figure}

\noindent{\textbf{Coordinate Descriptor Visualization.}}
To further explore how the feature volume describes the underlying structure, we visualize the corresponding coordinate descriptors queried in the feature volume.
Specifically, we accumulate the coordinate descriptors on each ray, resulting in a high-dimensional feature map.
PCA~\cite{pca} is utilized to reduce the dimension to 3 for visualization.
\cref{fig:featurevis} shows that the feature volume serves as a coarse structure template.
The face outline, hair, and background can be recognized easily.
Impressively, the eyes have a strong symmetry even with the glasses. 
Compared to raw coordinates, the feature descriptor provides a structured constraint to guide the image synthesis so that our method inherently synthesize image with better visual quality and 3D properties.

\begin{figure}[t]
    \centering
    \includegraphics[width=0.9\linewidth]{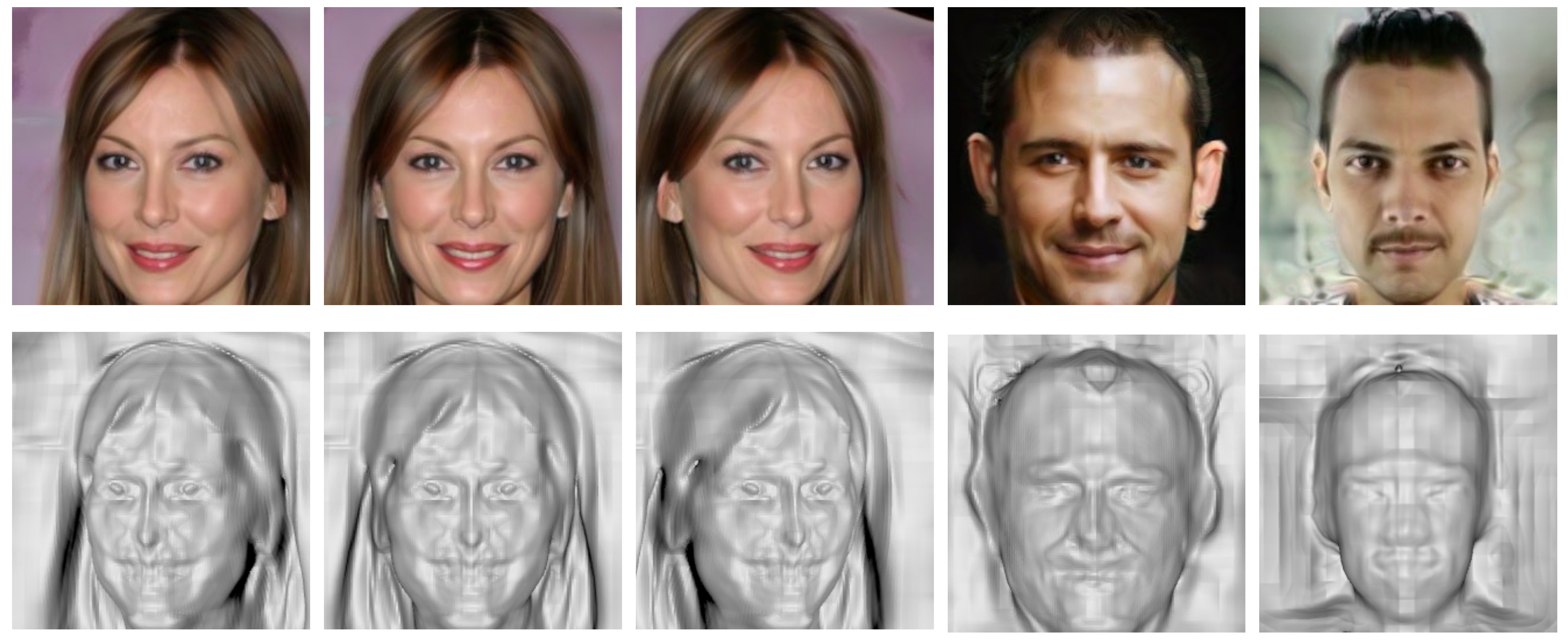}
    \vspace{-10pt}
    \caption{\textbf{3D Mesh extracted from the density.}}
   \vspace{-20pt}
    \label{fig:shape}
\end{figure}

\vspace{2pt}
\noindent{\textbf{Underlying Geometry.}}
The volume density of the implicit representation can construct an underlying geometry of the object due to its view-independent properties.
We extract the underlying geometry with marching cube~\cite{lorensen1987marching} on the density, resulting in a surface mesh.
\cref{fig:shape} shows the meshes with various views and identities.
The geometry is consistent across different views, supporting the good 3D properties of our method.

\section{Conclusion and Discussion}\label{sec:conclusion}

In this paper, we propose a new 3D-aware generative model, \textit{VolumeGAN}, for synthesizing high-fidelity images.
By learning structural and textural representations, our model achieves sufficiently higher image quality and better 3D control on various challenging datasets.

\noindent{\textbf{Limitations.}}
Despite the structural representation learned by VolumeGAN, the synthesized 3D mesh surface is still not smooth and lacks fine details.
Meanwhile, even though we can improve the synthesis resolution via introducing a deeper CNN (\textit{i.e.}, the neural renderer), it may weaken the multi-view consistency and 3D control.
Future research will focus on generating fine-grained 3D shape as well as making the tailing CNN in VolumeGAN with improved 3D properties through introducing regularizers.
%

\noindent{\textbf{Ethical Consideration.}}
Due to the high-quality 3D-aware synthesis performance, our approach is potentially applicable for deep fake generation. 
We strongly oppose the abuse of our method in violating privacy and security.
On the contrary, we hope it can be used to improve the existing fake detection systems.

\noindent{\textbf{Acknowledgement:}}  This work is supported in part by the Early Career Scheme (ECS) through the Research Grants Council (RGC) of Hong Kong under Grant No.24206219, CUHK FoE RSFS Grant, and Centre for Perceptual and Interactive Intelligence (CPII) Ltd under the Innovation and Technology Fund.

{\small
\bibliographystyle{ieee_fullname}
\bibliography{ref}
}

\end{document}


\title{3D-aware Image Synthesis via Learning Structural and Textural Representations \\ Supplementary Material}

\author{Yinghao Xu$^{1}$  \quad Sida Peng$^{2}$ \quad Ceyuan Yang$^{1}$ \quad Yujun Shen$^{3}$ \quad Bolei Zhou$^{1}$ \\
	$^1$The Chinese University of Hong Kong \quad
    $^2$Zhejiang University  \quad 
    $^3$Bytedance Inc. \\
    {\tt\small \{xy119, yc019, bzhou\}@ie.cuhk.edu.hk \quad    pengsida@zju.edu.cn} \quad 
	{\tt\small shenyujun0302@gmail.com} 
	}
	

\maketitle

\section{Overview}

This supplementary material is organized as follows.
\cref{sec:arch} and \cref{sec:config} introduce the network structure and the training configurations used in VolumeGAN.
\cref{sec:baseline-details} describes the details of implementing baseline approaches.
\cref{sec:additional-results} shows more qualitative results.
We also attach a demo video (\url{https://www.youtube.com/watch?v=p85TVGJBMFc}) to show the continuous 3D control achieved by our VolumeGAN.

\section{Network Structure}\label{sec:arch}

Recall that, our VolumeGAN first learns a feature volume with \textit{3D CNN}.
The feature volume is then transformed into a feature field using a \textit{ NeRF-like model}.
A 2D feature map is finally accumulated from the feature field and rendered to an image with a \textit{2D CNN}.
Taking 256 resolution as an instance, we illustrate the architectures of these three models in \cref{tab:3dcnn}, \cref{tab:2dmlp}, and \cref{tab:2dcnn}, respectively.
%

\begin{table}[b]
    \centering
    \caption{
        Network structure for learning a feature volume as the structural representation.
        The output size is with order $\{C \times H \times W \times D\}$, where $D$ denotes the depth dimension.
    }
    \label{tab:3dcnn}
    \vspace{-10pt}
    \begin{tabular}{ccccccc}
    \toprule
    Stage &  Block &   Output Size  \\ 
    \midrule
    input & Learnable Template &  $256\times 4 \times 4 \times 4$ \\
    \midrule
    \multirow{4}{*}{block$_1$} & \blocks{{128}}{{64}}{3} & \multirow{4}{*}{$128\times 8 \times 8 \times 8$} \\
    & &  \\
    & &  \\
     & &  \\
    \midrule
    \multirow{4}{*}{block$_2$} & \blocks{{64}}{{128}}{4}  &  \multirow{4}{*}{$64\times 16 \times 16 \times 16$} \\
    & &  \\
    & &  \\
     & &  \\
    \midrule
    \multirow{4}{*}{block$_3$} & \blocks{{32}}{{256}}{6} &
    \multirow{4}{*}{$32\times 32 \times 32 \times 32$} \\
    & &  \\
    & &  \\
     & &  \\
    \bottomrule
  \end{tabular}
\end{table}

\begin{table}[t]
    \centering
    \caption{
        Network structure of the generative feature field.
        The output size is with order $\{H \times W \times S \times C\}$, where $S$ is the number of sampling points along a certain camera ray.
        $\mathtt{FiLM}$ denotes the FiLM layer~\cite{perez2018film} and $\mathtt{Sine}$ stands for the Sine activation~\cite{siren}.
    }
    \label{tab:2dmlp}
    \vspace{-10pt}
    \begin{tabular}{ccccccc}
    \toprule
    Stage &  Block &   Output Size  \\ 
    \midrule
    input & $-$ &  $64 \times 64 \times12 \times(32+3)$ \\
    \midrule
    \multirow{3}{*}{mlp$_1$} & \blockb{{256}}{{64}}{3} & \multirow{3}{*}{$64 \times 64 \times12\times256$} \\
    & &  \\
    & &  \\
    \midrule
    \multirow{3}{*}{mlp$_2$} & \blockb{{256}}{{128}}{4}  &  \multirow{3}{*}{$64 \times 64 \times12\times256$} \\
    & &  \\
    & &  \\
    \midrule
    \multirow{3}{*}{mlp$_3$} & \blockb{{256}}{{256}}{6} &
    \multirow{3}{*}{$64 \times 64 \times12\times256$} \\
    & &  \\
    & &  \\
    \midrule
    \multirow{3}{*}{mlp$_4$} & \blockb{{256}}{{256}}{6} &
    \multirow{3}{*}{$64 \times 64 \times12\times256$} \\
    & &  \\
    & &  \\
    \bottomrule
  \end{tabular}
  \vspace{-25pt}
\end{table}

\begin{table}[b]
    \centering
    \caption{
        Network structure of the neural renderer, which renders a 2D feature map to a synthesized image.
        The output size is with order $\{C \times H \times W\}$.
    }
    \label{tab:2dcnn}
    \vspace{-10pt}
    \begin{tabular}{ccccccc}
    \toprule
    Stage &  Block &   Output Size  \\ 
    \midrule
    input & $-$ &  $256\times 64 \times 64$ \\
    \midrule
    \multirow{5}{*}{block$_1$} & \blocka{{128}}{{64}}{3} & \multirow{5}{*}{$128\times 128 \times 128$} \\
    & &  \\
    & &  \\
    & &  \\
    & &  \\
    \midrule
    \multirow{5}{*}{block$_2$} & \blocka{{64}}{{128}}{4}  &  \multirow{5}{*}{$64\times 256 \times 256$} \\
    & &  \\
    & &  \\
    & &  \\
    & &  \\
    \midrule
    RGB  & 3$\times$ 3 $\mathtt{Conv}$, 3 & $3\times 256 \times 256$ \\
    \bottomrule
  \end{tabular}
\end{table}

\begin{table*}[t]
    \centering
    \caption{
        Training configurations regarding different datasets.
    }
    \label{tab:config}
    \vspace{-5pt}
    \begin{tabular}{l|cccccccc}
    \toprule
    Datasets & Fov & Range$_{depth}$ & $\#$Steps & Range$_h$ & Range$_v$ & Sample\_Dist & $\lambda$ \\ \hline
    CelebA  & 12 & $[0.88, 1.12]$ & 12 & $[\pi/2-0.3, \pi/2+0.3]$ & $[\pi/2-0.15, \pi/2+0.15]$ & Gaussian & 0.2\\
    Cat    & 12 & $[0.8, 1.2]$ & 12 & $[\pi/2-0.5, \pi/2+0.5]$ & $[\pi/2-0.4, \pi/2+0.4]$ & Gaussian & 0.2  \\
    Carla  & 30 & $[0.7, 1.3]$ & 36 & $[0, 2\pi]$ & $[\pi/2-\pi/8, \pi/2+\pi/8]$ & Uniform & 1   \\
     FFHQ  & 12 & $[0.8, 1.2]$ & 14 & $[\pi/2-0.4, \pi/2+0.4]$ & $[\pi/2-0.2, \pi/2+0.2]$ & Gaussian & 1 \\
     CompCars  & 20 & $[0.8, 1.2]$ & 30 & $[0, 2\pi]$ & $[\pi/2-\pi/8, \pi/2+\pi/8]$ & Uniform & 1   \\
      Bedroom   & 26 & $[0.7, 1.3]$ & 40 & $[\pi/2-\pi/8, \pi/2+\pi/8]$ & $[\pi/2-\pi/10, \pi/2+\pi/10]$ & Uniform & 1  \\
    \bottomrule
    \end{tabular}
\end{table*}

\section{Training Configurations}\label{sec:config}

Because of the wildly divergent data distribution, the training parameters vary greatly on different datasets.
\cref{tab:config} illustrates the detailed training configuration of different datasets.
Fov, Range$_{depth}$, and $\#$Steps are the field of view, depth range and the number of sampling steps along a camera ray.
Range$_{h}$ and Range$_{v}$ denotes the horizontal and vertical angle range of the camera pose $\xi$.
'Sample\_Dist' denotes the sampling scheme of the camera pose. 
We only use Gaussian or Uniform sampling in our experiments.
$\lambda$ is the loss weight of the gradient penalty.

\section{Implementation Details of Baselines}\label{sec:baseline-details}

\noindent\textbf{HoloGAN~\cite{nguyen2019hologan}.}
We use the official implementation of HoloGAN.%
\footnote{\href{https://github.com/thunguyenphuoc/HoloGAN}{https://github.com/thunguyenphuoc/HoloGAN}}
We train HoloGAN for 50 epochs.
The generator of HoloGAN can only synthesize images in $64\times64$ or $128\times$ resolution.
We extend the generator with an extra $\mathtt{Upsample}$ and $\mathtt{AdaIN}$ block to synthesize $256\times256$ images for comparison.

\noindent\textbf{GRAF~\cite{graf}.}
We use the official implementation of GRAF.%
\footnote{\href{https://github.com/autonomousvision/graf}{https://github.com/autonomousvision/graf}}
We directly use the pre-trained checkpoints of CelebA and Carla provided by the authors.
For the other datasets, we train GRAF with the same data and camera parameters as ours at the target resolution.

\noindent\textbf{$\pi$-GAN~\cite{pigan}.}
We use the official implementation of $\pi$-GAN.%
\footnote{\href{https://github.com/marcoamonteiro/pi-GAN}{https://github.com/marcoamonteiro/pi-GAN}}
We also directly use the pre-trained checkpoints of CelebA, Carla and Cat for comparison and retrain $\pi$-GAN models on the other three datasets, including FFHQ, CompCars and LSUN bedroom.
The retrained models are progressively trained from a resolution of $32\times32$ to $256\times256$ following the official implementation.

\noindent\textbf{GIRAFFE~\cite{giraffe}.}
We use the official GIRAFFE implementation.%
\footnote{\href{https://github.com/autonomousvision/giraffe}{https://github.com/autonomousvision/giraffe}}
GIRAFFE provides the pre-trained weights of FFHQ and CompCars in a resolution of $256\times256$.
The remaining datasets are also trained with the same camera distribution for a fair comparison.

\section{Additional Results}\label{sec:additional-results}

\begin{figure*}[t]
    \centering
    \includegraphics[width=0.85\linewidth]{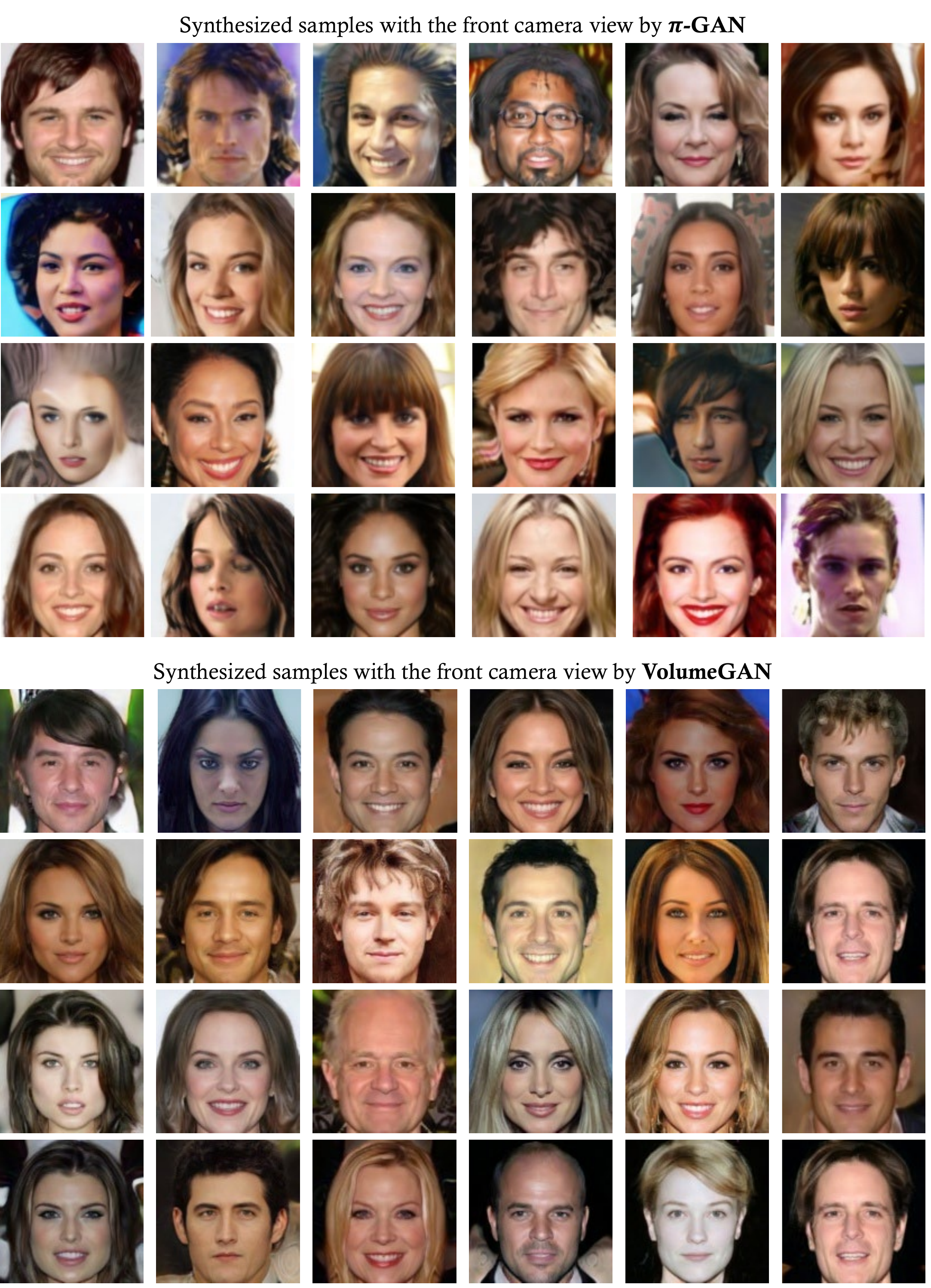}
    \caption{
        \textbf{More Synthesized results with the front camera view} by $\pi$-GAN~\cite{pigan} and our VolumeGAN, where the faces proposed by VolumeGAN are more consistent with the given view, suggesting a better 3D controllability.
    }
    \label{fig:more_pose_error}
\end{figure*}

\noindent\textbf{Synthesis with \textit{front camera view}.}
To better illustrate the 3D controllability, we show additional results of generating images with the front view.
As shown in \cref{fig:more_pose_error}, the faces synthesized by VolumeGAN are more consistent with the given view, demonstrating a better 3D controllability.

\vspace{5pt}
\noindent\textbf{Synthesis with \textit{varying camera views}.}
 Besides the front camera view, we also include a demo video (\url{https://www.youtube.com/watch?v=p85TVGJBMFc}), which shows more results with varying camera views.
From the video, we can see the continuous 3D control achieved by our VolumeGAN.
We also include comparisons with the state-of-the-art methods, \textit{i.e.}, $\pi$-GAN~\cite{pigan} and GIRRAFE~\cite{giraffe}, in the demo video.

{\small
\bibliographystyle{ieee_fullname}
\bibliography{ref}
}